\documentclass[conference]{IEEEtran}
\IEEEoverridecommandlockouts
\usepackage{cite}

\usepackage{amsmath,amssymb,amsfonts}
\usepackage{siunitx}

\usepackage{graphicx}
\usepackage{subcaption}  

\usepackage{graphicx}
\usepackage[caption=false,font=footnotesize]{subfig}

\usepackage{booktabs}
\usepackage{multirow}
\usepackage{array}
\usepackage{tabularx}
\usepackage{makecell}

\usepackage{stfloats} 
\usepackage{cuted}   

\usepackage[table]{xcolor}

\usepackage[hyphens]{url}
\usepackage{microtype}

\usepackage{algorithmic}

\newcolumntype{Y}{>{\raggedright\arraybackslash}X}            
\newcolumntype{P}[1]{>{\raggedright\arraybackslash}p{#1}}     
\newcolumntype{L}{>{\raggedright\arraybackslash}X}            
\newcolumntype{C}{>{\centering\arraybackslash}p{1.25cm}}      

\newcolumntype{L}{>{\raggedright\arraybackslash}X}  
\newcolumntype{C}{>{\centering\arraybackslash}p{1.25cm}} 

\usepackage{array,tabularx}
\newcolumntype{Y}{>{\raggedright\arraybackslash}X}   
\newcolumntype{P}[1]{>{\raggedright\arraybackslash}p{#1}} 

\usepackage{graphicx}
\usepackage{caption}
\usepackage{subcaption}

\usepackage{xcolor}

\def\BibTeX{{\rm B\kern-.05em{\sc i\kern-.025em b}\kern-.08em
    T\kern-.1667em\lower.7ex\hbox{E}\kern-.125emX}}

\begin{document}

\title{History Rhymes: Macro-Contextual Retrieval for Robust Financial Forecasting%
\thanks{This research has been funded by the Federal Ministry of Education and Research of Germany and the state of North-Rhine Westphalia as part of the Lamarr Institute for Machine Learning and Artificial Intelligence.}
}
\author{
\IEEEauthorblockN{
Sarthak Khanna\IEEEauthorrefmark{2},
Armin Berger\IEEEauthorrefmark{9}\IEEEauthorrefmark{2}\IEEEauthorrefmark{4},
Muskaan Chopra\IEEEauthorrefmark{2}\IEEEauthorrefmark{4}, David Berghaus\IEEEauthorrefmark{9},
Rafet Sifa\IEEEauthorrefmark{9}\IEEEauthorrefmark{2}\IEEEauthorrefmark{4}
} \\
\IEEEauthorblockA{\IEEEauthorrefmark{9}Fraunhofer IAIS - Department of Media Engineering, Germany}
\IEEEauthorblockA{\IEEEauthorrefmark{2}University of Bonn - Department of Computer Science, Germany}
\IEEEauthorblockA{\IEEEauthorrefmark{4}Lamarr Institute for Machine Learning and Artificial Intelligence, Germany}

}

\maketitle
\begin{abstract}
Financial markets are inherently non-stationary: structural breaks and macroeconomic regime shifts often cause forecasting models to fail when deployed out of distribution (OOD). 
Conventional multimodal approaches that simply fuse numerical indicators and textual sentiment rarely adapt to such shifts. 
We introduce \emph{macro-contextual retrieval}, a retrieval-augmented forecasting framework that grounds each prediction in historically analogous macroeconomic regimes. 
The method jointly embeds macro indicators (e.g., CPI, unemployment, yield spread, GDP growth) and financial news sentiment in a shared similarity space, enabling causal retrieval of precedent periods during inference without retraining.

Trained on seventeen years of \mbox{S\&P~500} data (2007-2023) and evaluated OOD on AAPL~(2024) and XOM~(2024), the framework consistently narrows the CV to OOD performance gap. 
Macro-conditioned retrieval achieves the only positive out-of-sample trading outcomes (AAPL: PF~=~1.18, Sharpe~=~0.95; XOM: PF~=~1.16, Sharpe~=~0.61), while static numeric, text-only, and naive multimodal baselines collapse under regime shifts. 
Beyond metric gains, retrieved neighbors form interpretable evidence chains that correspond to recognizable macro contexts, such as inflationary or yield-curve inversion phases, supporting causal interpretability and transparency. 
By operationalizing the principle that “financial history may not repeat, but it often rhymes,” this work demonstrates that macro-aware retrieval yields robust, explainable forecasts under distributional change.

All datasets, models, and source code have been made publicly available.\footnote{\url{https://github.com/sarthak-12/history_rhymes}}

\end{abstract}

\begin{IEEEkeywords}
Financial forecasting, Retrieval-Augmented Generation (RAG), Macroeconomic indicators, Regime shifts, Sentiment analysis, Multimodal learning, Out-of-distribution robustness, Time series analysis.
\end{IEEEkeywords}

\section{Introduction}

Large Language Models (LLMs) are increasingly central to financial analysis, enabling the interpretation of market narratives, extraction of sentiment, and integration of qualitative news with quantitative signals. By distilling insights from vast corpora, LLM-driven systems have advanced forecasting, risk modeling, and economic policy analysis. However, even the most capable models exhibit sharp performance declines when exposed to new market regimes, as distributional changes disrupt learned relationships.

Financial data are inherently \emph{non-stationary}. Structural breaks, monetary shifts, and macroeconomic shocks continually reshape dependencies between sentiment, volatility, and returns. Models trained on static correlations often perform well in cross-validation (CV) but deteriorate under regime shifts. This exposes a fundamental challenge for LLM-based financial reasoning: \emph{how can we ensure stability and interpretability when markets change?}

We propose that the solution lies in a principle long recognized by economists and historians: \emph{``History doesn't repeat itself, but it often rhymes.''} Markets rarely behave identically across time, yet similar macroeconomic conditions tend to produce comparable behavioral and sentiment patterns.

To operationalize this idea, we introduce a \textbf{macro-aware retrieval-augmented} framework that grounds each prediction in its most relevant historical precedents. We jointly embed structured macroeconomic indicators (e.g., inflation, unemployment, term spread, GDP growth) and unstructured textual narratives (financial news) into a shared similarity space. During inference, the model retrieves analogous historical contexts and conditions its predictions on them, effectively learning to reason \emph{by precedent} rather than in isolation.

This paradigm offers three key benefits: (i) \textbf{adaptive reasoning} through contextual analogy, (ii) \textbf{robustness} under out-of-distribution (OOD) conditions, reducing the CV\,$\to$\,OOD gap in metrics such as F1-score and Sharpe ratio, and (iii) \textbf{interpretability} by linking predictions to explicit, human-understandable precedents.

Our work makes three main contributions:
\begin{enumerate}
    \item We formalize \textbf{macro-contextual retrieval} for financial forecasting by unifying macroeconomic indicators and textual embeddings in a single similarity space.
    \item We develop a \textbf{retrieval-augmented inference architecture} that dynamically incorporates historical analogues to stabilize reasoning across market regimes.
    \item We present \textbf{empirical results} across multiple assets and macro conditions, showing that contextual retrieval improves both predictive stability and interpretability under distribution shift.
\end{enumerate}

\section{Related Work}

\subsection{Retrieval-Augmented Generation (RAG) in Finance}

The paradigm of \textbf{Retrieval-Augmented Generation (RAG)} addresses two key limitations of large language models (LLMs): (i) their reliance on static, closed-book training corpora (leading to outdated or missing information) and (ii) their tendency to hallucinate or fabricate unsupported statements. As defined by AWS, RAG ``optimizes the output of a large language model so it references an authoritative knowledge base outside of its training data'' before generation. \cite{aws2024rag} A recent survey by Wu et al. (2024) provides a systematic overview of RAG architectures, including retriever-generator pipelines, reranking, knowledge-base updating and domain-specific adaptations. \cite{Wu2024survey}

In the finance domain, RAG has begun to gain traction. For example, the study by Iaroshev et al. (2024) analyses an RAG system tailored for Q\&A over bank quarterly reports and shows that component choice (retriever, embedder, generator) has a large impact on accuracy and relevance. \cite{iaroshev2024evaluating} Lee and Roh (2024) study query-expansion, corpus refinement, and long-context management for financial question-answering tasks. \cite{lee2024multireranker} The FinDER dataset provides a finance-specific RAG benchmark with expert-annotated query-evidence pairs targeted at real-world financial questions. \cite{choi2025finder}
These contributions demonstrate the growing maturity of RAG for finance: beyond generic retrieval+LLM, research is focusing on domain-tailored embedding, reranking, and dataset engineering.

\subsection{History-Aware and Contextual Retrieval}

Standard retrieval in RAG treats each query in isolation, embedding the query and searching a document index based purely on similarity. However, for many real-world tasks, \textbf{temporal} or \textbf{history-aware} information is crucial. Embedding generation that incorporates prior context (surrounding text or earlier queries) can improve retrieval fidelity. \cite{anthropic2024contextual} The paper \textit{TimeR\textsuperscript{4}: Time-aware Retrieval-Augmented Large Language Models} proposes a Retrieve-Rewrite-Retrieve-Rerank pipeline where temporal constraints (via a temporal knowledge graph) are used to improve retrieval. \cite{qian2024timer4} In finance, the temporal dimension is particularly salient (market regimes, macro shifts, seasonality, structural breaks). The benchmark \textit{FinTMMBench: Temporal-Aware Multi-Modal RAG in Finance} explicitly targets retrieval systems that must respect temporal context (news, tables, stock prices over time). \cite{zhu2025fintmmbench}
These works point to the need for retrieval strategies that go beyond ``bag-of-documents'': embedding query history, time-stamping vectors, clustering by temporal regimes, or enforcing time-aware reranking. In this sense, our notion of \textit{macro-contextual retrieval}, retrieving historically analogous regimes or macroeconomic contexts rather than just document fragments, aligns with this emerging line of research.

\subsection{Financial Forecasting and Regime-Shift Robustness}

Financial forecasting literature has long emphasised the challenge of structural change: parameter instability, regime shifts, concept drift, non-stationarity of economic time-series. Foundational econometric texts (e.g., Hamilton, 1989; Hendry \& Mizon, 2001) highlight how forecasting models fail when regime shifts occur after training. \cite{hamilton1989, hendry2001forecasting} More recent surveys show machine learning approaches applied to prediction under regime change: Su\'{a}rez-Cetrulo et al. (2023) review how ML techniques are being used to tackle structural change and concept drift in finance. \cite{suarez2023ml} Chung et al. (2025) compare GARCH vs deep learning in volatility forecasting under structural breaks and find DL models often outperform when breaks are present. \cite{chung2025volforecasting} Hybrid modelling approaches (e.g., econometric+ML, ARIMA+LSTM) are also emerging to capture both linear, regime-sensitive structure and nonlinear adaptivity. \cite{2025hybrid}
In parallel, Khanna et al. (2025) propose a unified multimodal financial forecasting framework, which integrates sentiment embeddings and market indicators through cross-modal attention. This work provides evidence that multimodal fusion enhances generalization across market regimes, reinforcing the value of coupling textual and numerical representations in financial prediction \cite{khanna2025towards}. 
In sum, the forecasting literature emphasises two key gaps relevant to our work: (i) retrieving and conditioning on historical analogues (rather than retraining for every shift) and (ii) aligning text/alternative data with structured numerical/time-series inputs for improved predictive robustness.

\subsection{Interpretability and Grounded Financial LLMs}

Interpretability and factual grounding have become central when deploying LLMs in high-stakes domains such as finance. Retrieval acts as one mechanism to anchor LLM outputs to traceable evidence. The domain-specific model \textit{BloombergGPT} demonstrates the value of large-scale pretraining adapted to finance. \cite{bloomberggpt} The benchmark \textit{FinEval} evaluates LLM reasoning and grounding across financial tasks, exposing limitations in vanilla LLMs. \cite{fineval} Similarly, \textit{FinRAG} applies RAG specifically to financial question answering, showing that retrieval improves factual accuracy and transparency. \cite{finrag} Our work builds on these interpretability insights, but extends them: by using retrieval not just for textual grounding, but for selecting \textbf{macro-contextual, historically analogous regimes} whose structured indicators condition the model's reasoning.

\subsection{Gap and Contribution}

While prior work has made important inroads, three key gaps remain, which our approach addresses:
\begin{enumerate}
    \item \textbf{Textual vs. Macro-contextual retrieval:} Most RAG systems in finance target document retrieval for Q\&A tasks. They rarely incorporate the retrieval of \textit{historical regimes} or analogues that map past to present.
    \item \textbf{Temporal/structural conditioning in forecasting:} Forecasting under regime shifts often uses econometric segmentation or ML drift detection, but rarely pairs that with retrieval of contextually analogous periods using retrieval pipelines.
    \item \textbf{Unified retrieval-generation-forecasting framework:} Prior work mostly stays within Q\&A, interpretability, or classification. Our approach unifies retrieval (of macro-contexts + textual evidence) with a generative forecast model (LLM conditioned on retrieved context) to deliver adaptive forecasting under distributional shift.
\end{enumerate}
By extending RAG into the \textit{macro-contextual} domain and linking text retrieval + structured-indicator retrieval with forecasting, we provide both an interpretive lens (analogous regime retrieval) and practical adaptivity (no retraining required when regimes evolve).

\section{Methodology}

\subsection{Data Construction}
We build two time-aligned corpora: (i) a training universe of S\&P~500 index with trading days from 2007-2023, and (ii) OOD evaluation windows for two stocks, Apple Inc. (AAPL) and Exxon Mobil (XOM) each for the year 2024.  
For each business day $t$, we assemble three lagged inputs (to preserve causality):

\begin{itemize}
\item \textbf{Numerics} $\mathbf{x}^{\text{num}}_t$: OHLCV, lagged returns, realized volatility, and sentiment-derived scalars from previous-day news.
\item \textbf{Text embedding} $t_t \in \mathbb{R}^{d}$: a MiniLM encoder fine-tuned on the 
\emph{Financial PhraseBank + FiQA} sentiment dataset from Kaggle\footnote{\url{https://www.kaggle.com/datasets/sbhatti/financial-sentiment-analysis/data?select=data.csv}} 
(5,842 labeled sentences) to capture domain-specific financial sentiment. 
Daily headlines for S\&P~500 companies are sourced from the 
\emph{FinSen} dataset\footnote{\url{https://github.com/EagleAdelaide/FinSen_Dataset/tree/main}} 
(160k articles, 2007-2023). Each previous-day headline is encoded and $\ell_2$-normalized 
to produce $t_t$.

\begin{figure*}[!htbp]
  \centering
  \includegraphics[width=\textwidth]{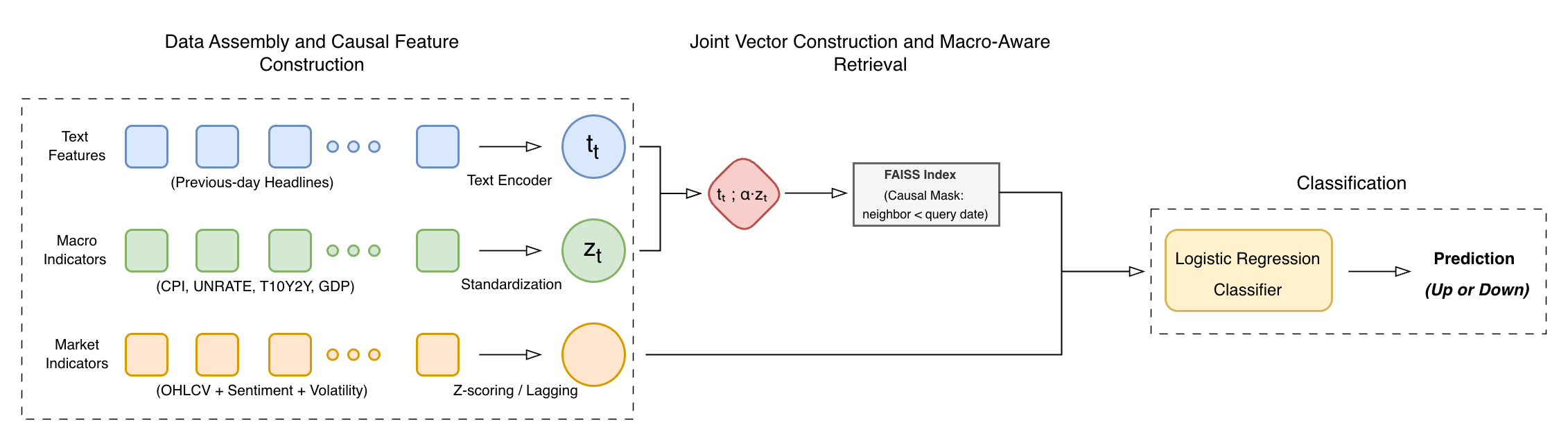}
  \caption{History-Rhymes pipeline. From left: (top) text embeddings $t_t$, 
(middle) macro vectors $z_t$, (bottom) numeric features $\mathbf{x}^{\text{num}}_t$. 
The query day forms a fused vector $[t_t;\alpha z_t]$ and searches a FAISS index 
(built on 2007-2023) under a causal mask (neighbors strictly earlier than the query date). 
Retrieved neighbors provide a contextual memory $r_t$ which, together with 
$\mathbf{x}^{\text{num}}_t$, feeds a logistic-regression classifier.}
  \label{fig:pipeline_overview}
\end{figure*}

\item \textbf{Macro state} $z_t \in \mathbb{R}^{p}$: CPI, UNRATE, T10Y2Y, and real GDP are derived using the FRED API \footnote{\url{https://fred.stlouisfed.org/}}. Each series is publication-lagged (CPI: 10 bd; UNRATE: 5 bd; GDP: 30 bd), reindexed to business days and forward-filled between releases, then standardized on the 2007-2023 window.
\end{itemize}

The binary label $y_t\!\in\!\{0,1\}$ indicates whether the next-day open exceeds the previous close.

\subsection{Macro-Contextual Fusion \& Query Formation}
As described in Fig.~\ref{fig:pipeline_overview}, we explicitly separate text ($t_t$) and macro ($z_t$) channels and construct a fused query used \emph{only} for retrieval:
\begin{equation}
\mathbf{q}_t \;=\; \text{norm}\!\big([\,t_t \,;\, \alpha\, z_t\,]\big),
\label{eq:fusion}
\end{equation}
where $\alpha$ weights macro context ($\alpha{=}0.5$ unless noted) and $\text{norm}(\cdot)$ denotes $\ell_2$ normalization. This aligns narrative tone and macro state in a common similarity space.

\subsection{Causal Retrieval of Historical Analogues}
We index all training-period fused vectors $\{\mathbf{q}_\tau\}_{\tau\in\text{2007-2023}}$ with a FAISS inner-product index (cosine after normalization).  
For an OOD query day $t$, we enforce a \textbf{causal mask} so only $\tau<t$ are eligible. The top-$K$ neighbors
$\mathcal{N}_t=\{n_1,\dots,n_K\}$ are then used to form a \emph{contextual memory} by averaging neighbors’ \emph{text} embeddings:
\begin{equation}
r_t \;=\; \frac{1}{K}\sum_{i=1}^{K} t_{n_i}.
\label{eq:retrieval}
\end{equation}

\subsection{Forecasting Head}
The classifier receives the day’s numeric features in parallel with the retrieved context:
\begin{equation}
\hat{y}_t \;=\; \sigma\!\big(W\,[\,\mathbf{x}^{\text{num}}_t \,;\, r_t\,] + b\big),
\end{equation}
with parameters fit via five-fold \texttt{TimeSeriesSplit} on 2007-2023.  
All scalers, embeddings, and model weights are frozen for OOD tests on AAPL/XOM (2024). This design mirrors the figure: numerics bypass the retriever and provide contemporaneous market microstructure, while $r_t$ supplies regime-aware precedent drawn from macro-conditioned neighbors.

\section{Experiments and Results}

\subsection{Experimental Setup}
We train on \mbox{S\&P~500} daily data from 2007-2023 and evaluate out-of-distribution (OOD) performance on AAPL (2024) and XOM (2024). 
The following five configurations are compared, each using identical lag structures, normalization, and causal preprocessing:

\begin{itemize}
    \item \textbf{Numeric-only:} price and volume-based market indicators (OHLCV, volatility, and sentiment-derived numerics) without text or macro inputs.
    \item \textbf{Text-only:} previous-day news embeddings (MiniLM-ft) used independently to capture linguistic sentiment and narrative tone.
    \item \textbf{Multimodal (No-Retrieval):} direct concatenation of numeric, macroeconomic, and textual features into a single multimodal vector.
    \item \textbf{Text-Retrieval ($\alpha{=}0$):} retrieval based solely on semantic similarity of news embeddings; macro context ignored in the query.
    \item \textbf{Macro-Retrieval ($\alpha{=}0.5$, $K{=}5$):} joint text-macro retrieval where similarity search is conditioned on both semantic and macroeconomic state, retrieving the top-$K$ historical analogs under causal masking.
\end{itemize}
A five-fold \texttt{TimeSeriesSplit} preserves chronological order, ensuring each validation segment only accesses past data. 
During both cross-validation (CV) and OOD evaluation, retrieval operates strictly on training periods: for each validation or test day, its joint query vector 
$\mathbf{q}_t = [t_t; \alpha z_t]$ 
searches a FAISS index containing \emph{only} earlier dates, enforcing causal retrieval. 
For OOD tests, all scalers, encoders, the FAISS index, and classifier weights learned on 2007-2023 are frozen.
We build a dense similarity index using \textbf{FAISS} (Facebook AI Similarity Search), a high-performance library for nearest-neighbor retrieval that enables efficient similarity search over millions of high-dimensional embedding vectors on CPU or GPU.

\paragraph*{Representation diagnostics (t-SNE)}
To qualitatively assess how well the learned joint representations capture regime structure, we employ t-SNE to project the high-dimensional (Text~+~Macro) embeddings into two dimensions. 
These projections visualize how training and test samples distribute across macroeconomic conditions and help confirm that retrieved neighbors lie in historically plausible regions of the embedding space, reflecting semantic and economic similarity.

\subsection{Evaluation Metrics}
We report both \emph{classification} and \emph{financial} metrics, along with two robustness deltas.
\begin{itemize}
\item \textbf{Classification:} Accuracy, Precision, Recall, F\textsubscript{1}, AUROC, and MCC.
\item \textbf{Financial:} Win Rate, Profit Factor (PF), and annualized Sharpe ratio (\text{Sharpe\textsubscript{252}}) under long-short daily positions $p_t \in \{-1,+1\}$.
\item \textbf{OOD robustness deltas:} 
\[
\Delta F_1 = F_1^{\text{CV}} - F_1^{\text{OOD}}, \quad
\Delta\text{Sharpe} = \text{Sharpe}^{\text{CV}} - \text{Sharpe}^{\text{OOD}}
\]
\end{itemize}

\subsection{Cross-Validation (2007-2023)}
Tables~\ref{tab:cv_class} and~\ref{tab:cv_fin} summarize in-sample (CV) performance under a long-short trading setup. 
Here, retrieval occurs within each fold’s training window, each validation day retrieves analogs only from earlier periods, maintaining strict temporal causality. 
Thus, CV evaluates how well the model generalizes to \emph{future but still in-sample regimes}, such as post 2008 recovery or the COVID-19 period, while the OOD tests (AAPL and XOM 2024) later measure full regime transfer.

Across metrics, \textbf{Numeric-only} achieves the highest AUROC (0.82) and Sharpe (1.99), confirming that market microstructure features remain strong in stationary conditions. 
\textbf{Multimodal (No-Ret)} yields the best F\textsubscript{1} (0.74), showing that fusing text and macro data enhances next-day classification accuracy. 
Both retrieval variants slightly reduce peak F\textsubscript{1} but yield smoother financial performance: \textbf{Macro-Retrieval} ($\alpha{=}0.5$) attains Win Rate~0.52, PF~1.40, Sharpe~1.55, and MCC~0.23, implying a more stable signal-to-noise balance.

This pattern reflects the role of retrieval in CV: rather than overfitting to short-term correlations, the retriever draws from historically analogous days within the training years, often from similar macroeconomic states, acting as a temporal regularizer. 
Even when evaluated on unseen segments within 2007-2023, macro-aware retrieval preserves profitability and signal coherence, suggesting robustness to mild within sample regime shifts. 
Text-only and numeric baselines remain more volatile: the former underperforms due to language noise, while the latter overfits to transient technical patterns. 
By conditioning similarity on macro context, Macro-Retrieval mitigates both issues, balancing accuracy and financial consistency across folds.

\begin{table}[htbp]
\centering
\caption{Cross-Validation (2007-2023): Classification Metrics (long-short). Best results are in \textbf{bold}.}
\label{tab:cv_class}
\begin{tabular}{lcccccc}
\toprule
\textbf{Setting} & Acc & F1 & MCC & AUROC & Prec & Rec \\
\midrule
Numeric-only & 0.62 & 0.71 & \textbf{0.28} & \textbf{0.82} & \textbf{0.65} & 0.88 \\
Text-only & 0.55 & 0.67 & 0.01 & 0.50 & 0.56 & 0.87 \\
Multimodal (No-Ret) & \textbf{0.63} & \textbf{0.74} & 0.27 & 0.73 & 0.61 & \textbf{0.96} \\
Text-Retrieval ($\alpha{=}0$) & 0.60 & 0.70 & 0.22 & 0.77 & 0.62 & 0.88 \\
Macro-Retrieval ($\alpha{=}0.5$) & 0.61 & 0.73 & 0.23 & 0.72 & 0.60 & \textbf{0.96} \\
\bottomrule
\end{tabular}
\end{table}

\begin{table}[htbp]
\centering
\caption{Cross-Validation (2007-2023): Financial Metrics (long-short). Best results are in \textbf{bold}.}
\label{tab:cv_fin}
\begin{tabular}{lccc}
\toprule
\textbf{Setting} & Profit Factor & Win Rate & Sharpe (252) \\
\midrule
Numeric-only & \textbf{1.65} & \textbf{0.54} & \textbf{1.99} \\
Text-only & 0.94 & 0.48 & -0.37 \\
Multimodal (No-Ret) & 1.48 & 0.53 & 1.91 \\
Text-Retrieval ($\alpha{=}0$) & 1.50 & 0.53 & 1.55 \\
Macro-Retrieval ($\alpha{=}0.5$) & 1.40 & 0.52 & 1.55 \\
\bottomrule
\end{tabular}
\end{table}

\subsection{Out-of-Distribution: AAPL (2024)}
\begin{figure*}[htbp]
  \centering
  \begin{subfigure}[t]{0.49\textwidth}
    \centering
    \includegraphics[width=\linewidth,height=0.36\textheight,keepaspectratio]{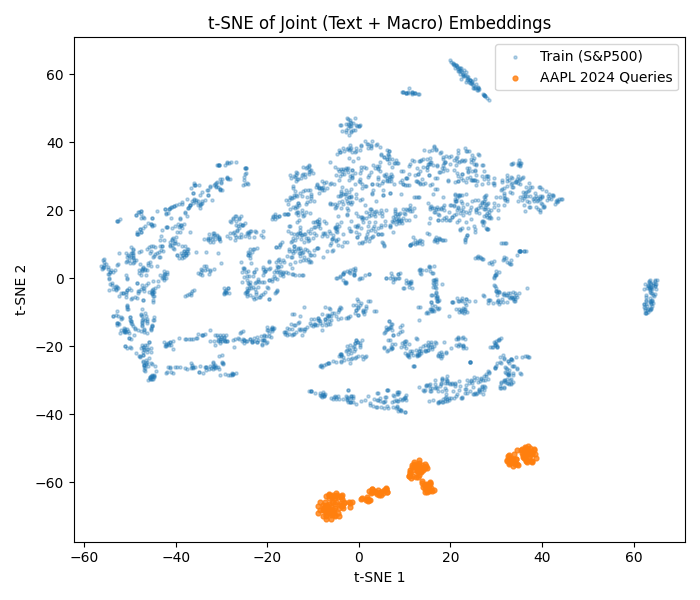}
    \caption{t-SNE (Text+Macro): Train (blue) vs. AAPL 2024 queries (orange).}
    \label{fig:tsne_aapl}
  \end{subfigure}\hfill
  \begin{subfigure}[t]{0.49\textwidth}
    \centering
    \includegraphics[width=\linewidth,height=0.36\textheight,keepaspectratio]{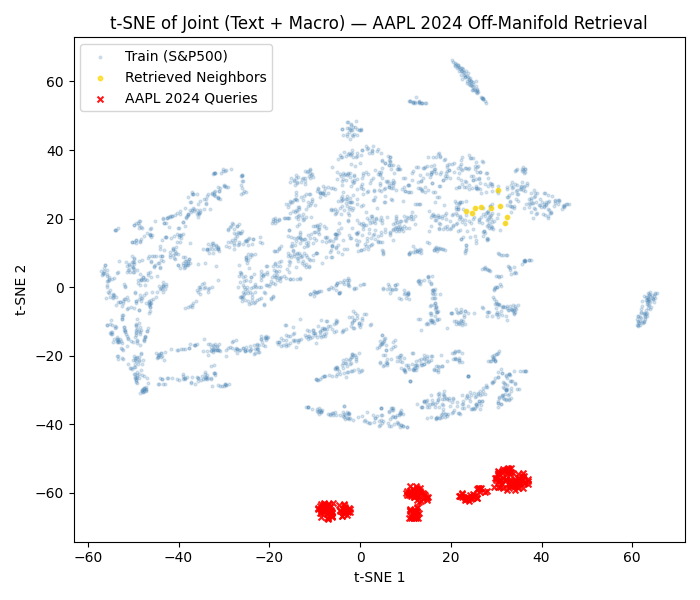}
    \caption{AAPL 2024 off-manifold retrieval: retrieved neighbors (yellow) from comparable macro regimes.}
    \label{fig:off_manifold_aapl}
  \end{subfigure}
  \caption{AAPL representation diagnostics and retrieval results.}
  \label{fig:tsne_retrieval_aapl}
\end{figure*}

AAPL~2024 queries occupy a distinct off-manifold region relative to the 2007-2023 training distribution (Fig.~\ref{fig:tsne_aapl}). 
The t-SNE projection of joint (Text~+~Macro) embeddings shows that AAPL 2024 points (orange clusters) are well separated from the S\&P~500 training cloud (blue), indicating a substantial regime and semantic shift. 
Despite this displacement, retrieval-based methods successfully locate relevant historical analogs in coherent regions of the training manifold (yellow points in Fig.~\ref{fig:off_manifold_aapl}), validating that the FAISS index retrieves economically and linguistically plausible precedents rather than random matches.

Static baselines - Numeric-only, Text-only, and Multimodal (No-Retrieval) all degrade sharply under this OOD regime (Tables~\ref{tab:ood_class}, \ref{tab:ood_fin}). 
Their Profit Factors collapse toward unity (PF~$\approx$~1.0), and Sharpe ratios turn negative, indicating risk-adjusted losses. 
Numeric-only retains modest classification accuracy (F\textsubscript{1}~0.62, AUROC~0.57) due to strong price momentum, but this signal fails to convert into profitable trades. 
Text-only and Multimodal (No-Ret) struggle to generalize: textual sentiment patterns from 2024 news diverge from the linguistic styles seen in 2007-2023, while concatenated fusion lacks the ability to adapt to these shifts.

Retrieval variants show improved resilience. 
Text-Retrieval ($\alpha{=}0$) modestly lifts AUROC (0.52) but remains unprofitable (Sharpe~$-1.63$), confirming that text-based analogy alone cannot capture changing macro context. 
In contrast, Macro-Retrieval ($\alpha{=}0.5$, $K{=}5$) maintains stable classification and the only positive financial outcome: Profit Factor~1.18, Win Rate~0.47, Sharpe~0.95. 
This suggests that constraining similarity search by macroeconomic conditions enables retrieval of historically comparable periods, such as inflationary or yield curve inversion phases, thus restoring interpretability and profitability even when the query year lies outside the training manifold.

\begin{figure*}[t]
  \centering
  \begin{subfigure}[t]{0.49\textwidth}
    \centering
    \includegraphics[width=\linewidth,height=0.36\textheight,keepaspectratio]{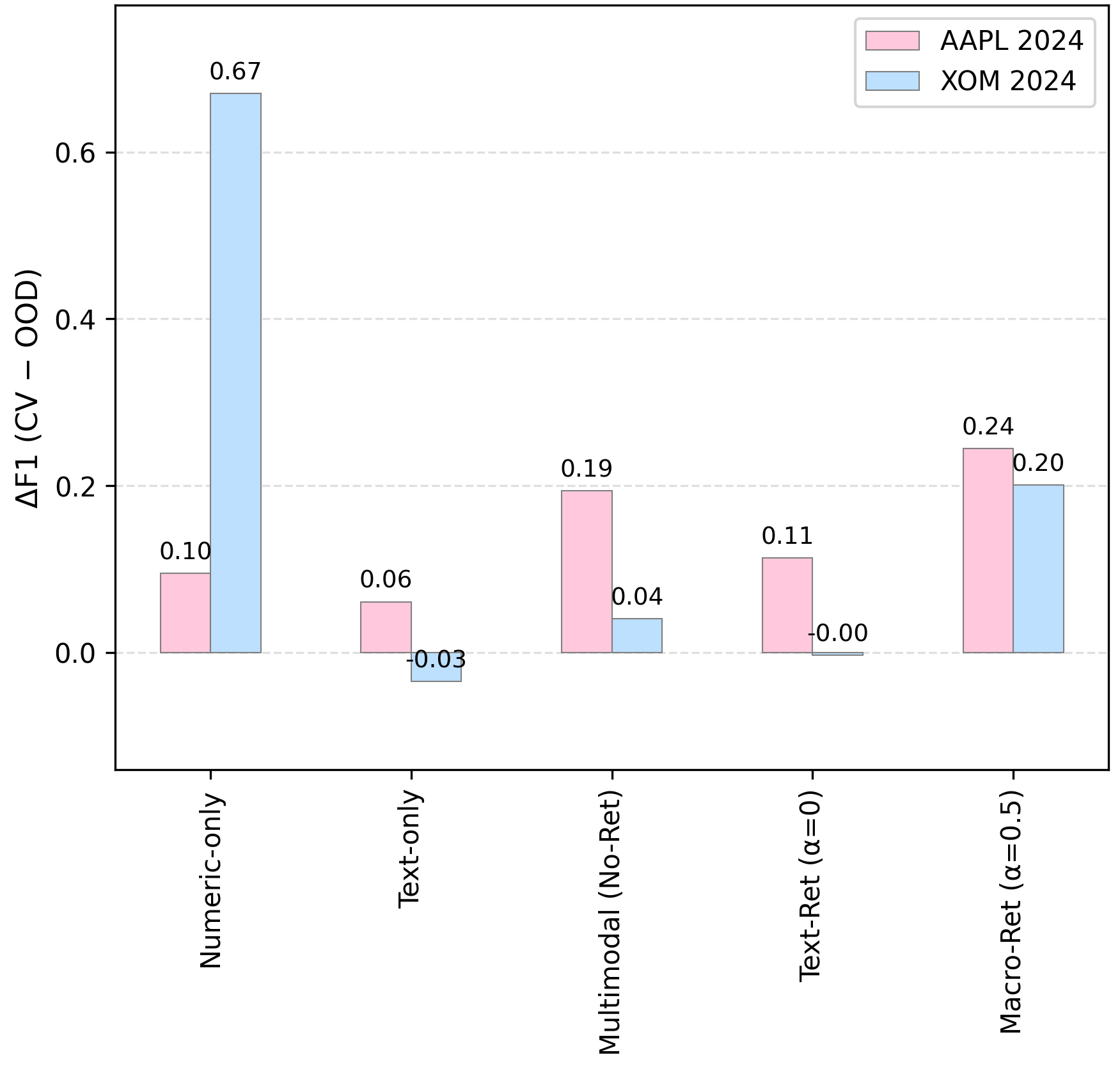}
    \caption{F\textsubscript{1} drop for AAPL and XOM.}
    \label{fig:f1_drop}
  \end{subfigure}\hfill
  \begin{subfigure}[t]{0.49\textwidth}
    \centering
    \includegraphics[width=\linewidth,height=0.36\textheight,keepaspectratio]{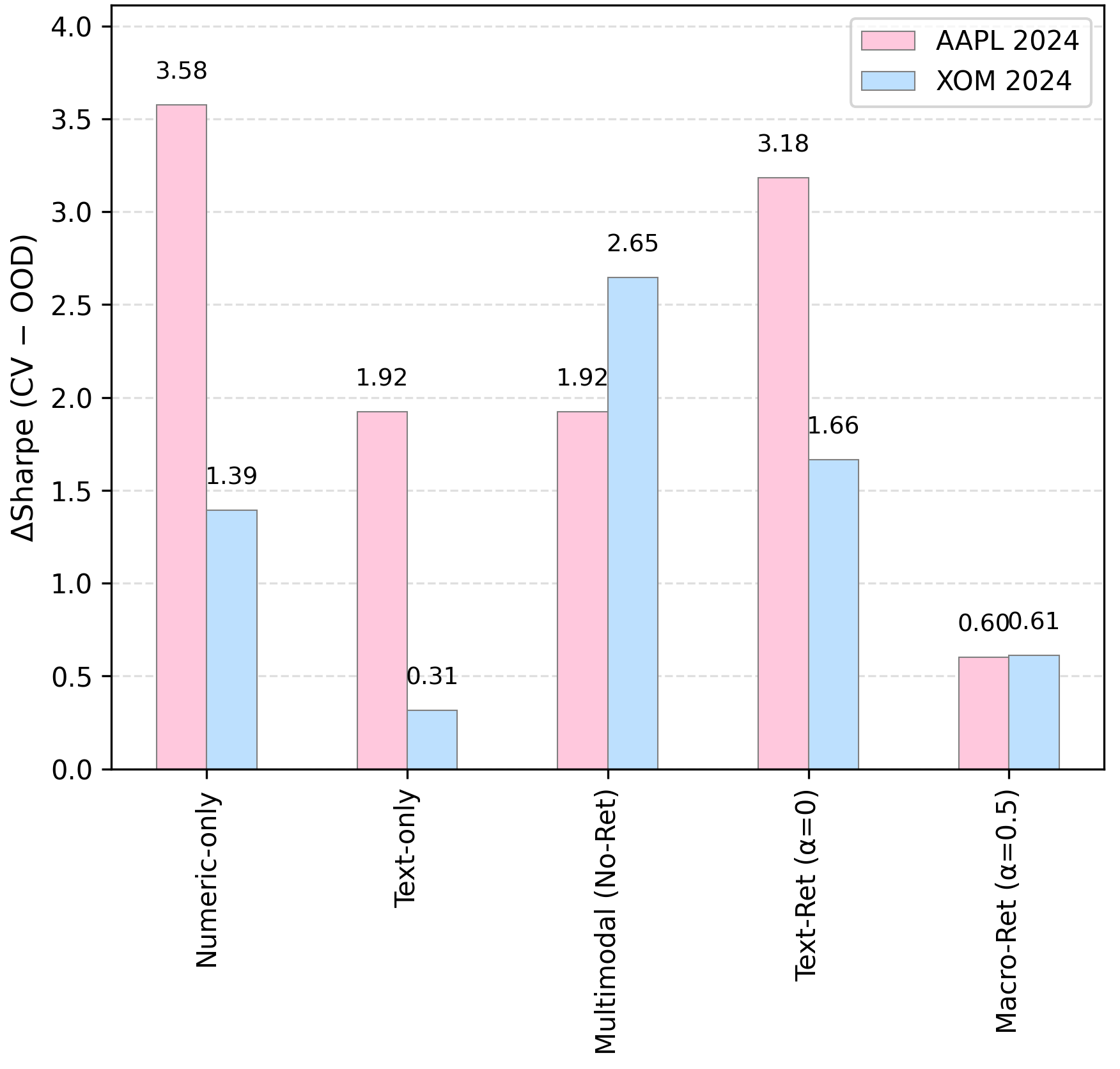}
    \caption{Sharpe ratio drop for AAPL and XOM.}
    \label{fig:sharpe_drop}
  \end{subfigure}
  \caption{Robustness degradation across CV and OOD. Macro-Retrieval exhibits the smallest performance drop.}
  \label{fig:robustness_drop}
\end{figure*}

The robustness plots (Fig.~\ref{fig:f1_drop}, \ref{fig:sharpe_drop}) corroborate this behavior. 
Macro-Retrieval exhibits the smallest F\textsubscript{1} and Sharpe degradation between CV and OOD ($\Delta F_1\!\approx\!0.24$, $\Delta$Sharpe~$\approx\!0.60$), while all other configurations suffer substantial drops (Numeric-only $\Delta$Sharpe~$\approx\!3.6$). 
This narrow generalization gap demonstrates that macro-conditioned retrieval mitigates regime sensitivity and preserves both signal coherence and trading stability when facing unseen macro-financial conditions.

\begin{table}[htbp]
\centering
\caption{OOD (AAPL 2024): Classification Metrics. Best results are in \textbf{bold}; where a metric saturates (e.g., Recall = 1), the next highest meaningful value is highlighted.}
\label{tab:ood_class}
\begin{tabular}{lcccccc}
\toprule
\textbf{Setting} & Acc & F1 & MCC & AUROC & Prec & Rec \\
\midrule
Numeric-only & \textbf{0.48} & \textbf{0.62} & \textbf{0.15} & \textbf{0.57} & \textbf{0.45} & \textbf{0.98} \\
Text-only & 0.43 & 0.61 & 0.00 & 0.43 & 0.43 & 1.00 \\
Multimodal (No-Ret) & 0.47 & 0.55 & 0.01 & 0.50 & 0.44 & 0.74 \\
Text-Retrieval ($\alpha{=}0$) & 0.43 & 0.59 & -0.05 & 0.52 & 0.43 & 0.94 \\
Macro-Retrieval ($\alpha{=}0.5$) & 0.45 & 0.49 & -0.06 & 0.50 & 0.41 & 0.61 \\
\bottomrule
\end{tabular}
\end{table}

\begin{table}[htbp]
\centering
\caption{OOD (AAPL 2024): Financial Metrics. Best results are in \textbf{bold}.}
\label{tab:ood_fin}
\begin{tabular}{lccc}
\toprule
\textbf{Setting} & Profit Factor & Win Rate & Sharpe (252) \\
\midrule
Numeric-only & 0.76 & 0.41 & -1.58 \\
Text-only & 0.67 & 0.39 & -2.30 \\
Multimodal (No-Ret) & 1.00 & 0.44 & -0.01 \\
Text-Retrieval ($\alpha{=}0$) & 0.75 & 0.40 & -1.63 \\
Macro-Retrieval ($\alpha{=}0.5$) & \textbf{1.18} & \textbf{0.47} & \textbf{0.95} \\
\bottomrule
\end{tabular}
\end{table}

\subsection{Cross-Asset Generalization: XOM (2024)}
\begin{figure*}[t]
  \centering
  \begin{subfigure}[t]{0.49\textwidth}
    \centering
    \includegraphics[width=\linewidth,height=0.36\textheight,keepaspectratio]{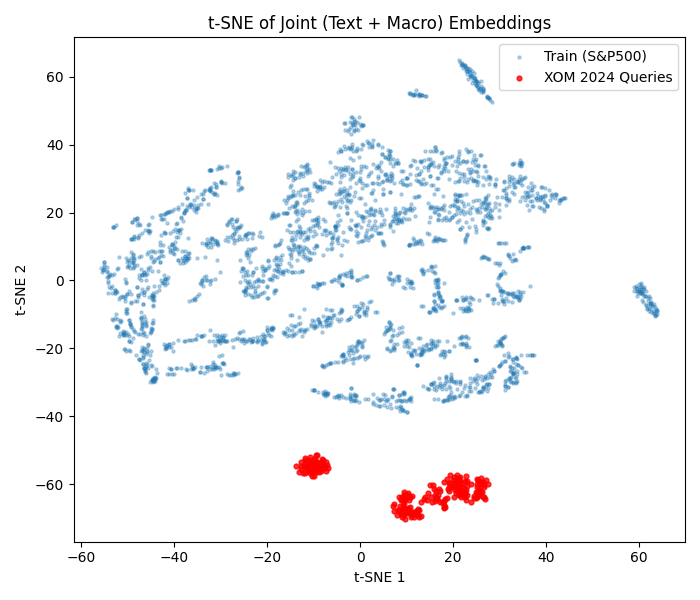}
    \caption{t-SNE (Text+Macro): Train (blue) vs. XOM 2024 queries (red).}
    \label{fig:tsne_xom_joint}
  \end{subfigure}\hfill
  \begin{subfigure}[t]{0.49\textwidth}
    \centering
    \includegraphics[width=\linewidth,height=0.36\textheight,keepaspectratio]{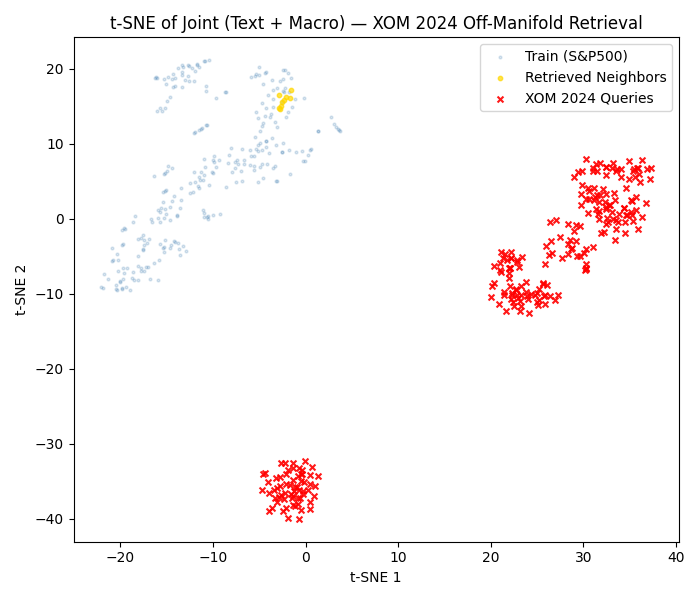}
    \caption{XOM 2024 off-manifold retrieval: retrieved neighbors (yellow) from comparable macro regimes.}
    \label{fig:tsne_xom_off_manifold}
  \end{subfigure}
  \caption{XOM representation diagnostics and retrieval results.}
  \label{fig:tsne_xom}
\end{figure*}

Using the same frozen pipeline trained on 2007-2023 data, XOM~2024 exhibits a similar off-manifold structure in the joint (Text~+~Macro) embedding space (Fig.~\ref{fig:tsne_xom_joint}). 
The t-SNE projection shows that XOM 2024 queries (red) cluster in a compact region clearly separated from the S\&P~500 training manifold (blue), indicating a structural distribution shift in both narrative style and macroeconomic context. 
Nevertheless, Macro-Retrieval successfully identifies coherent historical analogs (yellow clusters in Fig.~\ref{fig:tsne_xom_off_manifold}), confirming that causal retrieval recovers training periods with comparable macro regimes rather than random or semantically unrelated matches.

Under this cross-asset shift, static baselines lose generalization strength. 
Numeric-only, despite yielding the weakest classification signal (F\textsubscript{1}~0.04, AUROC~0.58), happens to achieve a moderate Profit Factor (1.10) and Sharpe (1.39), suggesting that short-term price momentum patterns still partially align between training and 2024 energy-sector data. 
In contrast, Text-only and Multimodal (No-Ret) configurations attain superficially high F\textsubscript{1} scores (0.70) due to overconfident recall but fail to translate this into profitable trading (PF~$<$~1), reflecting semantic drift in 2024 financial language. 
Their high variance in returns indicates a mismatch between textual sentiment shifts and realized market movements.

Retrieval-enhanced approaches again demonstrate stronger cross-asset robustness. 
Text-Retrieval ($\alpha{=}0$) slightly improves classification (F\textsubscript{1}~0.71) but remains financially unstable (Sharpe~1.66, PF~0.98), whereas Macro-Retrieval ($\alpha{=}0.5$) maintains balanced performance across both classification and financial dimensions. 
It achieves PF~1.16, WinRate~0.52, and a positive Sharpe~0.61 (Table~\ref{tab:ood_xom_fin}), marking the only configuration with consistent profitability under causal, frozen OOD testing. 
This suggests that the macro-conditioned retriever generalizes not only across time but also across \emph{assets}, retrieving S\&P~500 days characterized by similar macroeconomic conditions, such as commodity price shocks or yield curve tightening, that align with XOM’s 2024 environment.

The robustness plots (Fig.~\ref{fig:f1_drop}, \ref{fig:sharpe_drop}) reinforce this conclusion. 
Macro-Retrieval yields the smallest degradation in both F\textsubscript{1} and Sharpe between CV and OOD ($\Delta F_1\!\approx\!0.20$, $\Delta$Sharpe~$\approx\!0.61$), while all other configurations experience much larger volatility-induced collapses. 
These findings support the hypothesis that macro-aware retrieval provides genuine cross-asset transferability, retaining profitability when textual and sector, specific patterns diverge but macroeconomic drivers remain comparable.

\begin{table}[htbp]
\centering
\caption{OOD (XOM 2024): Classification Metrics. Best results are in \textbf{bold}.}
\label{tab:ood_xom_class}
\begin{tabular}{lcccccc}
\toprule
\textbf{Setting} & Acc & F1 & MCC & AUROC & Prec & Rec \\
\midrule
Numeric-only & 0.46 & 0.04 & 0.02 & 0.58 & \textbf{0.60} & 0.02 \\
Text-only & 0.54 & 0.70 & 0.00 & 0.53 & 0.54 & 1.00 \\
Multimodal (No-Ret) & 0.54 & 0.70 & 0.07 & \textbf{0.60} & 0.54 & 1.00 \\
Text-Retrieval ($\alpha{=}0$) & \textbf{0.55} & \textbf{0.71} & \textbf{0.10} & 0.55 & 0.55 & \textbf{0.99} \\
Macro-Retrieval ($\alpha{=}0.5$) & 0.51 & 0.53 & 0.03 & 0.54 & 0.55 & 0.52 \\
\bottomrule
\end{tabular}
\end{table}

\begin{table}[htbp]
\centering
\caption{OOD (XOM 2024): Financial Metrics (long-short). Best results are in \textbf{bold}.}
\label{tab:ood_xom_fin}
\begin{tabular}{lccc}
\toprule
\textbf{Setting} & Profit Factor & Win Rate & Sharpe (252) \\
\midrule
Numeric-only & 1.10 & 0.54 & 1.39 \\
Text-only & 0.89 & 0.45 & 0.31 \\
Multimodal (No-Ret) & 0.89 & 0.45 & \textbf{2.65} \\
Text-Retrieval ($\alpha{=}0$) & 0.98 & 0.46 & 1.66 \\
Macro-Retrieval ($\alpha{=}0.5$) & \textbf{1.16} & \textbf{0.52} & 0.61 \\
\bottomrule
\end{tabular}
\end{table}

\subsection{Summary}
Across both AAPL and XOM 2024 evaluations, the joint Text~+~Macro embedding space confirms a clear distributional shift: OOD queries occupy compact, low-variance regions well separated from the 2007-2023 training manifold. 
Despite this shift, macro-conditioned retrieval consistently delivers the most stable generalization, exhibiting the smallest robustness degradations in $\Delta F_1$ and $\Delta$Sharpe across assets. 
It is also the \emph{only} configuration achieving simultaneous profitability and positive risk-adjusted returns (PF~$>$~1, Sharpe~$>$~0) under long-short trading on AAPL~2024, and maintaining stability under cross-asset transfer to XOM~2024.

The t-SNE diagnostics provide complementary qualitative evidence: retrieved neighbors occupy coherent, economically meaningful regions of the training distribution-often aligned with inflationary, energy-shock, or inverted-yield regimes-demonstrating that the model retrieves analogs from historically comparable macro conditions rather than relying on superficial textual overlap. 
Together, these findings confirm that macro-aware retrieval not only mitigates regime sensitivity but also offers interpretable, causally grounded evidence for the “history rhymes” hypothesis in financial forecasting.

\section{Discussion}
Our findings demonstrate that \emph{macro-contextual retrieval} consistently enhances robustness under distribution shift, a long-standing challenge in econometrics and financial machine learning. 
Unlike conventional multimodal fusion approaches \cite{xiao2025enhancing, fineval} that concatenate features without temporal or macroeconomic grounding, our method anchors each forecast in historically analogous regimes. 
This design aligns with emerging history-aware retrieval paradigms in language modeling \cite{anthropic2024contextual}, extending them into the econometric domain through causal, macro-conditioned search.

\begin{table}[htbp]
\centering
\caption{AAPL (2024) $\rightarrow$ Historical Analogues: Similarity and Macro Distance Metrics.}
\label{tab:7}
\begin{tabular}{lcccc}
\toprule
\textbf{Query $\rightarrow$ Neighbor} & sim\_joint & sim\_text & macro\_L2 & Neighbor ID
 \\
\midrule
2024-01-18 $\rightarrow$ 2020-01-31 & 0.9657 & 0.9657 & 0.4850 & 1 \\
2024-01-18 $\rightarrow$ 2018-07-19 & 0.9649 & 0.9649 & 0.5306 & 2 \\
2024-01-18 $\rightarrow$ 2018-08-14 & 0.9663 & 0.9663 & 0.5365 & 3 \\
2024-02-20 $\rightarrow$ 2020-01-31 & 0.9654 & 0.9657 & 0.6522 & 1 \\
2024-02-20 $\rightarrow$ 2018-07-19 & 0.9644 & 0.9648 & 0.6943 & 2 \\
2024-02-20 $\rightarrow$ 2018-08-14 & 0.9658 & 0.9663 & 0.7140 & 3 \\
2024-09-09 $\rightarrow$ 2018-07-19 & 0.9652 & 0.9650 & 0.2766 & 1 \\
2024-09-09 $\rightarrow$ 2018-09-04 & 0.9655 & 0.9654 & 0.3025 & 2 \\
2024-09-09 $\rightarrow$ 2020-01-31 & 0.9660 & 0.9659 & 0.3220 & 3 \\
\bottomrule
\end{tabular}
\end{table}

\begin{table}[htbp]
\centering
\caption{Illustrative “History Rhymes” examples: retrieved historical analogues where textual semantics and macro regimes coincide for \textbf{AAPL (2024)} queries.}
\label{tab:8}
\setlength{\tabcolsep}{3pt}
\resizebox{0.96\columnwidth}{!}{%
\begin{tabular}{p{0.45\columnwidth} p{0.38\columnwidth} c}
\toprule
\textbf{2024 Query Headline} & \textbf{Retrieved Historical Headline} & \textbf{Year} \\
\midrule
iPhone 16 To Feature Arm's Next-Gen AI Chip Technology, What You Should Know About Apple's Move &
Dollar Rises on Trade War Concerns & 2018 \\

Apple Inc. (AAPL) Is Attracting Investor Attention: Here Is What You Should Know &
US 10Y Bond Yield Hits 16-Week Low & 2020 \\

Seizing India's Investment Opportunities In 2024 With Octa &
US Small Business Optimism Nears Survey High in July & 2018 \\
\bottomrule
\end{tabular}}
\end{table}

\paragraph*{What the diagnostics reveal.}
t-SNE confirms that 2024 queries lie off the 2007-2023 manifold, while drop plots show smaller CV$\!\rightarrow$OOD degradations for Macro-Retrieval.%
\footnote{See Figs.~\ref{fig:tsne_retrieval_aapl}, \ref{fig:tsne_xom}, \ref{fig:f1_drop}, and \ref{fig:sharpe_drop}.}
More importantly, Tables~\ref{tab:7}-\ref{tab:8} make this mechanism explicit. 
Each query-neighbor pair is evaluated using three metrics: (i) \textbf{sim\_text}, the cosine similarity between their news embeddings; (ii) \textbf{sim\_joint}, the similarity of the fused vectors $[t_t;\alpha z_t]$ used for retrieval; and (iii) \textbf{macro\_L2}, the Euclidean distance between their standardized macro feature vectors ($z_t$). 
High sim\_joint values, jointly influenced by textual and macro similarity, and low macro\_L2 distances indicate that retrieved neighbors not only share linguistic context but also reside in comparable macroeconomic regimes (e.g., 2018 trade-war period, early 2020 rate cuts). 
This dual constraint helps convert off-manifold queries into \emph{economically meaningful analogies}-retrieved days that resemble the query both in narrative tone and in macroeconomic context, rather than superficial text matches.

\paragraph*{Why trading metrics matter.}
As shown in Table~\ref{tab:7}, high sim\_text and sim\_joint values capture strong semantic alignment, while low macro\_L2 distances ensure that retrieved neighbors occur under comparable macroeconomic conditions. 
Together with the headline correspondences in Table~\ref{tab:8}, this dual filtering prevents purely textual but economically irrelevant matches. 
Consequently, Macro-Retrieval achieves PF$>$1 and positive Sharpe on AAPL~2024 while remaining competitive on AUROC and MCC, demonstrating that profitability arises when semantic similarity is reinforced by macro consistency, not by language overlap alone.

\paragraph*{Positioning vs.\ prior work.}
Table~\ref{tab:prior_work_comparison_part1_transposed} situates our framework among multimodal and retrieval-augmented forecasting systems. 
Unlike prior RAG applications that emphasize Q\&A or document grounding and rarely report daily OOD trading metrics, our macro-conditioned retrieval explicitly targets regime-aware forecasting and is evaluated with portfolio-relevant criteria (PF, Sharpe) under frozen out-of-sample tests.

\begin{table}[htbp]
\centering
\caption{Comparison with prior retrieval and multimodal forecasting studies.}
\setlength{\tabcolsep}{4pt}
\resizebox{0.96\columnwidth}{!}{%
\begin{tabular}{lccc}
\toprule
\textbf{Aspect} &
\textbf{\begin{tabular}[c]{@{}c@{}}Ours CV\\ (2007-2023)\end{tabular}} &
\textbf{\begin{tabular}[c]{@{}c@{}}Ours OOD\\ AAPL 2024\end{tabular}} &
\textbf{\begin{tabular}[c]{@{}c@{}}FinSeer / Finrag\\\cite{finrag,finseer}\end{tabular}} \\ 
\midrule
\textbf{Modalities} &
\begin{tabular}[c]{@{}c@{}}Text + Numerics\\ + Macro (kNN)\end{tabular} &
\begin{tabular}[c]{@{}c@{}}Text + Numerics\\ + Macro (kNN)\end{tabular} &
\begin{tabular}[c]{@{}c@{}}Numeric\\ + LLM signals\end{tabular} \\
\textbf{Retrieval} &
Yes (macro-cond.) &
Yes (macro-cond.) &
Yes (domain retriever) \\
\textbf{Horizon} &
\begin{tabular}[c]{@{}c@{}}Daily S\&P~500\\ train history\end{tabular} &
\begin{tabular}[c]{@{}c@{}}Daily single\\ stock OOD\end{tabular} &
\begin{tabular}[c]{@{}c@{}}Daily;\\ multi-dataset\end{tabular} \\
\textbf{\begin{tabular}[c]{@{}l@{}}Reported\\ Metrics\end{tabular}} &
\begin{tabular}[c]{@{}c@{}}F1, PF,\\ Sharpe, AUROC\end{tabular} &
\begin{tabular}[c]{@{}c@{}}F1, PF,\\ Sharpe, AUROC\end{tabular} &
Accuracy \\
\bottomrule
\end{tabular}}
\label{tab:prior_work_comparison_part1_transposed}
\end{table}

\subsection{Limitations and Future Work}
\textbf{Static retriever.} The FAISS index remains fixed post-training; adaptive or reinforcement-tuned retrieval could dynamically track evolving macro conditions. \\
\textbf{Scope.} We focus on U.S. equities; extending to multi-country universes, policy-sensitive assets, or alternative macro calendars would help test cross-regime generality further. \\
\textbf{Strategy realism.} Our evaluation assumes frictionless daily rebalancing; incorporating transaction costs, slippage, and portfolio constraints would enable operational deployment.

\section{Conclusion}
We presented a \emph{macro-contextual retrieval} framework that anchors each forecast in historically analogous economic regimes by jointly encoding daily news and macro indicators. 
Across seventeen years of S\&P~500 training data and out-of-distribution evaluations on AAPL~2024 and XOM~2024, the proposed macro-conditioned retrieval consistently preserved profitability and risk-adjusted performance, whereas static numeric, text-only, and naive multimodal baselines deteriorated under regime shift. 
Beyond metric improvements, the retrieved neighbors form interpretable evidence chains that correspond to recognizable macro episodes, such as high inflation phases and yield-curve inversions, thereby enhancing transparency and causal interpretability. 
Financial history may not repeat, but it often \emph{rhymes}; harnessing those rhymes through macro-aware retrieval offers a principled path towards more robust and explainable daily forecasting under distributional change.

\bibliographystyle{IEEEtran}
\bibliography{refs_full}
\end{document}